\documentclass[10pt,twocolumn,letterpaper]{article}
\usepackage{xspace} 
\usepackage[table]{xcolor}

\usepackage{cvpr}              
\usepackage[numbers,sort&compress]{natbib}
\usepackage{multirow} 
\usepackage{tabularx}
\usepackage{dblfloatfix}
\usepackage{subcaption}
\usepackage[table]{xcolor}
\definecolor{lightgreen}{rgb}{0.9, 1, 0.9}
\usepackage{colortbl}
\definecolor{lightgreen}{rgb}{0.9, 1, 0.9}
\usepackage{flushend}
\usepackage{graphicx}
\usepackage{caption}
\usepackage{subcaption}
\definecolor{cvprblue}{rgb}{0.21,0.49,0.74}
\usepackage[pagebackref,breaklinks,colorlinks,citecolor=cvprblue]{hyperref}

\def\paperID{3663} 
\def\confName{CVPR}
\def\confYear{2024}

\title{MR-MLLM: Mutual Reinforcement of Multimodal Comprehension and \\Vision Perception}

\author{
    {\normalsize Guanqun Wang$^{1*}$ \quad Xinyu Wei$^{1*}$ \quad Jiaming Liu$^{1*}$ \quad Ray Zhang$^{2*}$ \quad Yichi Zhang$^{1}$}\\ 
    {\normalsize Kevin Zhang$^{1}$ \quad Maurice Chong$^{1}$ \quad Shanghang Zhang$^{1\dag}$}\\
    {\normalsize{$^{1}$National Key Laboratory for Multimedia Information Processing, School of Computer Science,}}\\
    {\normalsize Peking University \quad $^{2}$Shanghai AI Lab}
}

\begin{document}
\maketitle
\renewcommand{\thefootnote}{\fnsymbol{footnote}}
\footnotetext[1]{These authors contributed equally to this work.}
\footnotetext[2]{Corresponding authors: \href{mailto:shanghang@pku.edu.cn}{shanghang@pku.edu.cn}}

\def\thefootnote{\arabic{footnote}}
\begin{abstract}
In recent years, multimodal large language models (MLLMs) have shown remarkable capabilities in tasks like visual question answering and common sense reasoning, while visual perception models have made significant strides in perception tasks, such as detection and segmentation. 
However, MLLMs mainly focus on high-level image-text interpretations and struggle with fine-grained visual understanding, 
and vision perception models usually suffer from open-world distribution shifts due to their limited model capacity.
To overcome these challenges, we propose the Mutually Reinforced Multimodal Large Language Model (MR-MLLM), a novel framework that synergistically enhances visual perception and multimodal comprehension. 
First, a shared query fusion mechanism is proposed to harmonize detailed visual inputs from vision models with the linguistic depth of language models, enhancing multimodal comprehension and vision perception synergistically. Second, we propose the perception-enhanced cross-modal integration method, incorporating novel modalities from vision perception outputs, like object detection bounding boxes, to capture subtle visual elements, thus enriching the understanding of both visual and textual data. In addition, an innovative perception-embedded prompt generation mechanism is proposed to embed perceptual information into the language model’s prompts, aligning the responses contextually and perceptually for a more accurate multimodal interpretation.
Extensive experiments demonstrate MR-MLLM's superior performance in various multimodal comprehension and vision perception tasks, particularly those requiring corner case vision perception and fine-grained language comprehension. 
\end{abstract}

\section{Introduction}
\label{sec:intro}

In the pursuit of artificial intelligence that mirrors the intricacy of human cognition and perception, research on vision-language Multimodal Large Language Model (MLLM) has become a frontier of paramount significance~\citep{radford2021learning,li2023blip,fu2023challenger}. 
Recent studies in MLLM have been directed towards the joint processing of visual and textual modalities. 

\begin{figure}[h!]
    \centering
    \includegraphics[width=0.48\textwidth]{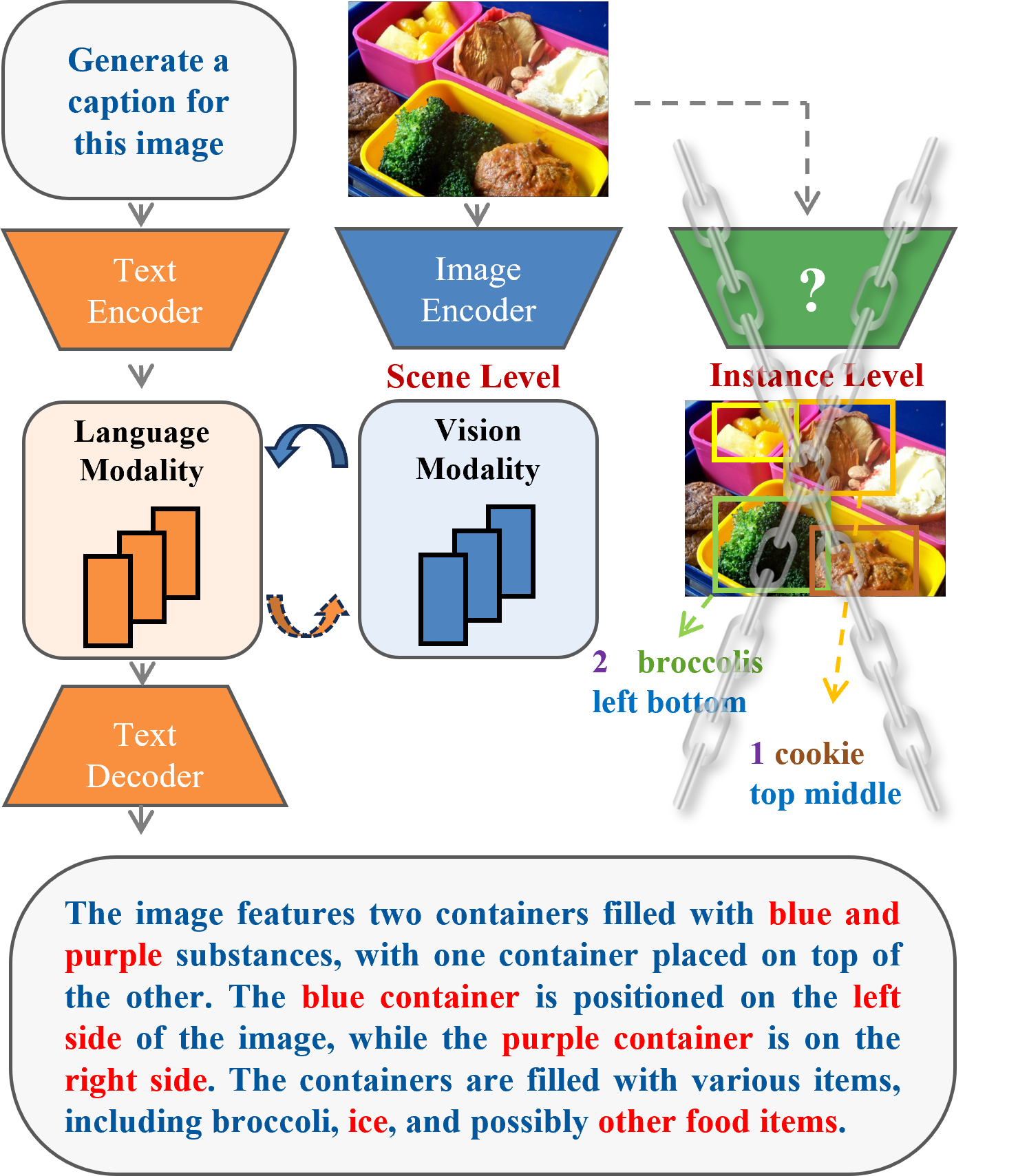}
    \caption{\textbf{Limitations of current MLLMs.} (1) Limited interaction between two modalities. (2) Acquisition of instance-level descriptors. (3) The enhancement of visual tasks using linguistic knowledge remains under-explored. The caption above is generated from LLaMA-Adapter V2~\cite{gao2023llamaadapter}. It fails to perceive fine-grained details within the images.}
    \label{fig:1a}
\end{figure}

However, MLLMs are primarily geared towards high-level image-text interpretations, yet they often encounter difficulties in fine-grained visual comprehension.
As shown in Figure \ref{fig:1a}, the predominant approach~\citep{li2023blip,li2023blip2,liu2023visual,chen2023shikra} involves the adoption of a CLIP~\citep{radford2021learning}-like contrastive learning paradigm, training image encoders and text encoders in a manner that aligns visual features with textual features. 
Such methods, while effective in feature alignment, often lack in facilitating deeper interactions between modalities, consequently perpetuating the chasm between them.
Furthermore, although some efforts~\citep{alayrac2022flamingo,zhang2023llamaadapter, gao2023llamaadapter} have engaged in the direct computation of visual and textual features, enhancing the interplay between the two modalities, the image-text pretrained visual encoders (e.g., ViT-L from CLIP, Qformer~\citep{li2023blip}) tend to yield only high-level scene descriptors of images. 

\begin{figure}[t]
    \centering
    \includegraphics[width=0.46\textwidth]{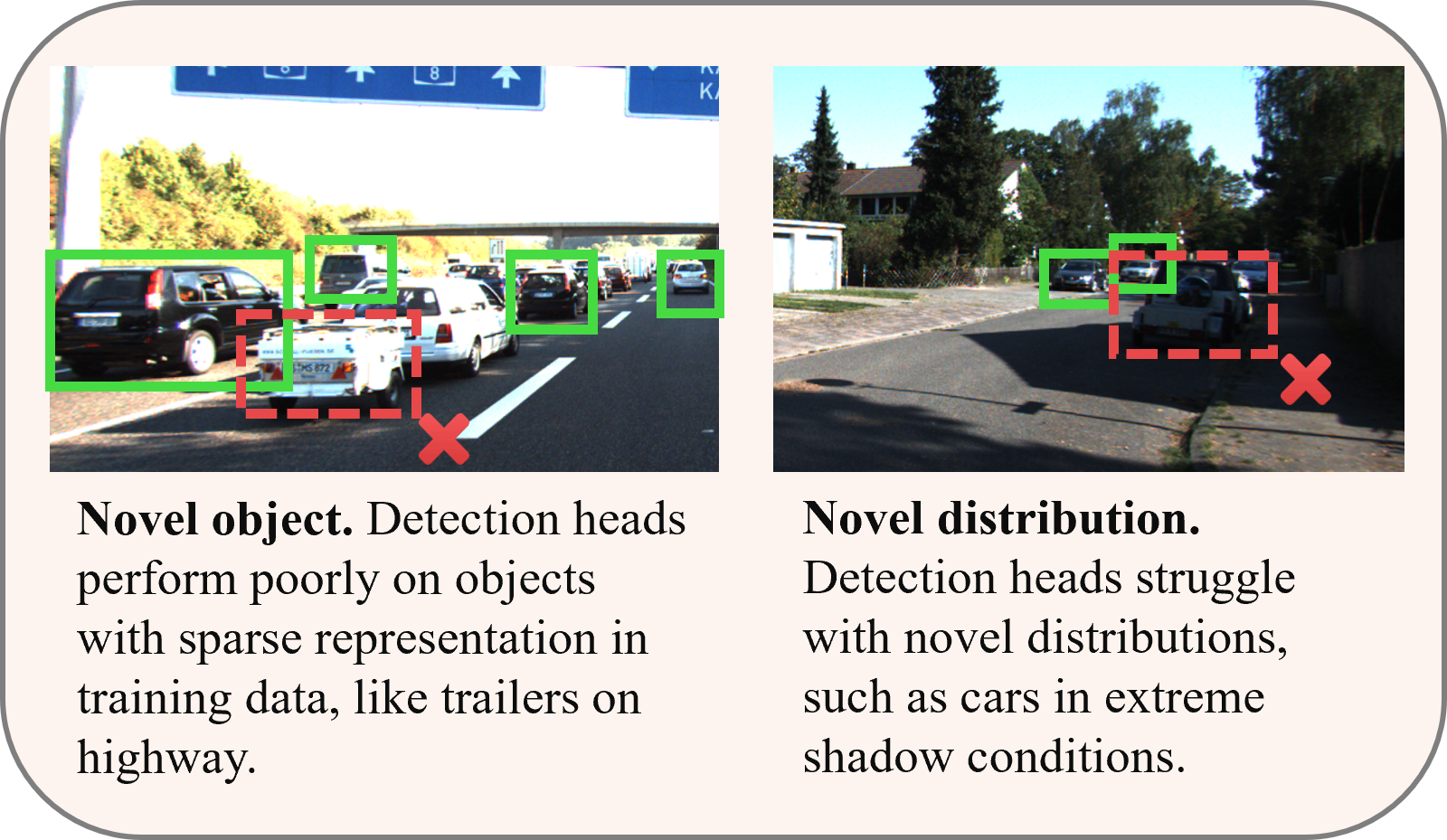}
    \caption{\textbf{Limitations of perception models in Conercases.} Detection outcomes above stem from DETR~\citep{DETR}. It lacks generalization capabilities and world knowledge in scene interpretation.}
    \label{fig:1b}
\end{figure}

In the meanwhile, vision perception models, often trained within specific domains, exhibit limitations in their generalization capabilities, as shown in Figure \ref{fig:1b}, especially when confronted with open-world scenarios~\citep{carion2020end,liu2023grounding,zong2023detrs}. 
The ability to accurately perceive and interpret corner cases, which are frequently encountered in real-world settings, remains a critical issue. 
The prevailing methods, despite being effective within their training contexts, struggle to maintain performance when exposed to the diverse and unpredictable nature of real-world data.

In this paper, we propose to integrate the powerful generalization and emergent abilities of MLLM, with the fine-grained perception capabilities of vision perception models, to overcome the limitations of each other.
To this end, we design the Mutually Reinforced Multimodal Large Language Model (MR-MLLM), a framework designed to synergistically enhance both vision perception and multimodal understanding. Our approach incorporates innovative mechanisms to deepen the interplay between visual and linguistic modalities:
First, we introduce a shared query fusion mechanism in MR-MLLM. This mechanism harmonizes detailed visual inputs from vision models with the linguistic depth of language models, thereby enriching multimodal comprehension and enhancing vision perception synergistically.
Second, our model adopts a perception-enhanced cross-modal integration approach. By incorporating novel modalities from vision perception outputs, such as object detection bounding boxes, MR-MLLM captures and integrates subtle visual elements, enriching the understanding of both visual and textual data.
\begin{figure*}[t] 
\centering 
\includegraphics[width=\textwidth]{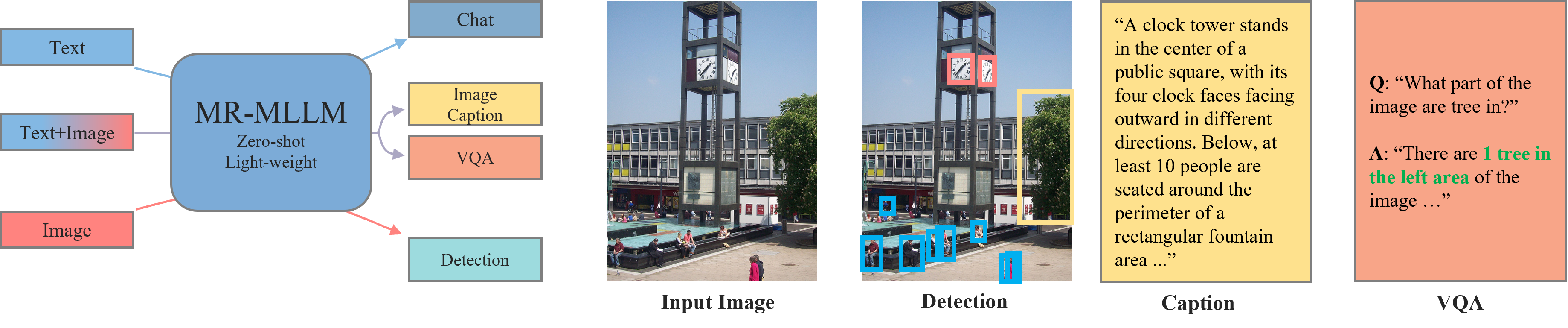} 
\caption{\textbf{Capabilities of MR-MLLM. } Through unified-format visual instruction tuning, MR-MLLM is capable of a range of vision-language tasks like Image Captioning and Visual Question Answering (VQA). It refines the detection outcomes of object detection heads, leveraging the world knowledge and generalization capabilities of LLMs to assist vision tasks. Concurrently, this refinement process also enhances the model's fine-grained perception abilities. } 
\label{Fig1} 
\end{figure*}
Additionally, we have developed an innovative perception-embedded prompt generation method. This method embeds perceptual information into the language model’s prompts, aligning the responses contextually and perceptually, which leads to more accurate and comprehensive multimodal interpretation.The range of tasks that our method can handle is illustrated in Figure \ref{Fig1}.

Extensive experiments are conducted on several visual question answering, image captioning, and object detection benchmarks. It shows that our MR-MLLM surpasses the state-of-the-art MLLMs, such as BLIP2~\citep{li2023blip}, LLaMA-Adapter V2~\citep{gao2023llama}, Shikra~\citep{chen2023shikra}, Instruct BLIP~\citep{Dai2023InstructBLIPTG}, and Qwen-VL~\citep{Bai2023QwenVLAF}, in the fine-grained multimodal comprehension tasks. In particular, MR-MLLM achieves 71.5\% accuracy on the visual space reasoning benchmark, VSR~\citep{Liu2022VisualSR}, which far exceeds that of other MLLMs with the same size of parameters. In addition, our MR-MLLM significantly enhances the detection capabilities of vision perception models in handling corner case detection tasks. On the CODA detection dataset~\citep{li2022coda}, MR-MLLM boosts the corner case average recall of the baseline vision perception model by 2\%, and endows closed-set trained specialized detectors with the ability to detect novel classes.
In summary, experimental results demonstrate MR-MLLM's superior performance in a range of multimodal comprehension and vision perception tasks, particularly those requiring fine-grained visual understanding and sophisticated language comprehension.
Our contributions can be summarized as follows:
\begin{itemize}
\item We propose the Mutually Reinforced Multimodal Large Language Model (MR-MLLM), designed to synergistically combine the generalization and emergent capabilities of MLLM with the fine-grained perception abilities of vision perception models, thereby achieving mutual enhancement across modalities.
\item We introduce a novel shared query fusion mechanism in MR-MLLM, enhancing the interplay between detailed visual perception and language comprehension.
\item We propose a perception-enhanced cross-modal integration method, integrating novel modalities to capture subtle visual details for a richer understanding of multimodal content.
\item We develop a perception-embedded prompt generation approach, ensuring contextually and perceptually coherent responses in multimodal interpretations.
\item Through extensive experiments, we demonstrate the effectiveness of MR-MLLM in managing intricate multimodal comprehension and advanced vision perception challenges. These findings suggest that MR-MLLM represents a significant advancement in the field of multimodal learning, contributing to ongoing efforts to enhance the integration of visual and linguistic data.
\end{itemize}

\section{Related Work}
\label{sec:related_work}
\textbf{Multimodal Large Language Model.} 
 CLIP~\citep{radford2021learning} pioneered training both text and image encoders for feature alignment across modalities, defining the MLLM concept. Subsequent models like Flamingo~\citep{alayrac2022flamingo} and BLIP2~\citep{li2023blip2} advanced this integration, with methods ranging from cross-attention layers to modality connection via Q-Former. LLaMA-Adapter~\citep{zhang2023llamaadapter,gao2023llama}, FROMAGe~\citep{koh2023grounding}, and LLaVA~\citep{liu2023visual} simplified this by directly feeding visual features into LLMs. Our approach also directly integrates visual features but focuses on fine-grained object descriptors from detection outputs to enhance LLMs, maintaining a lower parameter count. 
\\
\textbf{Object Detection.} After the introduction of the transformer-based end-to-end object detection model DETR~\citep{DETR}, numerous studies~\citep{group-detr, dynamic-head, fastc-detr, hy-detr, fasttc-detr, anchor-detr, D-detr,zhang2022monodetr} have advanced the field, continually setting new state-of-the-art (SOTA) benchmarks across various object detection datasets. Following the emphasis on query de-noising in lightweight models like DN-DETR and DINO, which achieved robust detection performance, the next focal point in object detection has shifted towards open-set detection capabilities, i.e., the ability to detect categories not seen during training. Grounding-DINO, by integrating the Transformer-based detector DINO with grounded pre-training, demonstrates the potential to detect arbitrary objects using human inputs such as category names or referring expressions.

Our approach aims to leverage the world knowledge inherent in Large Language Models (LLMs) to enhance object detection models' performance in corner cases while refining open-set detection models' ability to distinguish between categories.
\\
\textbf{Position Representation.} The representation of specific regions within images has long been an active area of research. DisClip~\citep{bracha2023disclip} directly inputs cropped image patches with the original image into model. Several studies like~\citep{MMIIS} emphasize particular areas by using binary masks or Gaussian maps as input. SAM~\citep{kirillov2023segment} encodes prompt points or prompt boxes into positional embeddings. Some models are required to output coordinates. Traditional anchor-based methods use sliding windows and proposal candidate regions to locate the vertices of bounding boxes, like Fast R-CNN~\citep{fastrcnn}. Anchor-free methods such as FCOS~\citep{tian2019fcos} regress the four coordinates of bounding boxes directly to eliminate anchors. Transformer-based object detection methods, like DETR~\citep{DETR}, convert object detection into an end-to-end manner using one-to-one label assignments.
\\
In the realm of LLMs, VisionLLM~\citep{wang2023visionllm} directly uses linguistic representations for coordinates but alters the language model's vocab. Shikra~\citep{chen2023shikra} and Sphinx~\citep{lin2023sphinx} directly use language for representing coordinates. We also employ language directly for coordinate representation without altering the vocab. Moreover, we introduce a transformer-based detection head, whose queries also implicitly express coordinate information.
\\
\textbf{Instruction Tuning.} Instruction tuning, as shown in InstructGPT~\citep{instructGTP}, FLAN~\citep{FLAN}, and others~\citep{OPT-IML,guo2023point,han2023imagebind,yang2023lidarllm}, significantly boosts LLMs' performance. Inspired by this, Flamingo~\citep{alayrac2022flamingo} utilizes visual and linguistic inputs as prompts and achieves impressive zero-shot results. LLaMA-Adapter~\citep{zhang2023llama}, LLaVA~\citep{llava} and MINIGPT4~\citep{zhu2023minigpt} employ generated visual instruction-following data. However, these studies primarily focus on image-to-text tasks without exploring how LLMs can facilitate visual tasks. VisionLLM~\citep{wang2023visionllm} and Shikra~\citep{chen2023shikra} aim to align visual and linguistic tasks by using language instructions to solve both types of tasks uniformly, prompting the model to directly output bounding box coordinates. However, due to the inherent differences between images and language, an LLM cannot outperform a specialized object detection model. To mitigate this, we introduce detection head into MLLM and design object detection data in an instruction-tuning format, enabling the model to refine coordinates. We also discover that this process reciprocally enhances the LLM's capability in language tasks.

\begin{figure*}[ht]
  \centering
  \begin{subfigure}[b]{\textwidth}
    \centering
    \includegraphics[width=\textwidth]{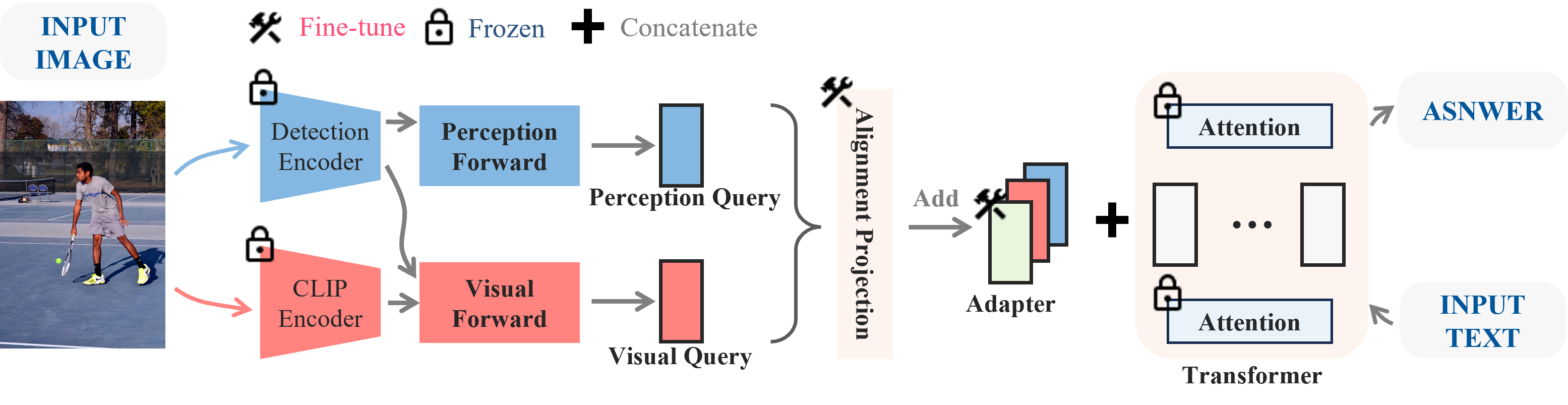}
    \caption{\textbf{General Pipeline.} }
    \label{fig:imagea}
  \end{subfigure}

  \begin{subfigure}[b]{0.5\textwidth}
    \centering
    \includegraphics[width=\linewidth]{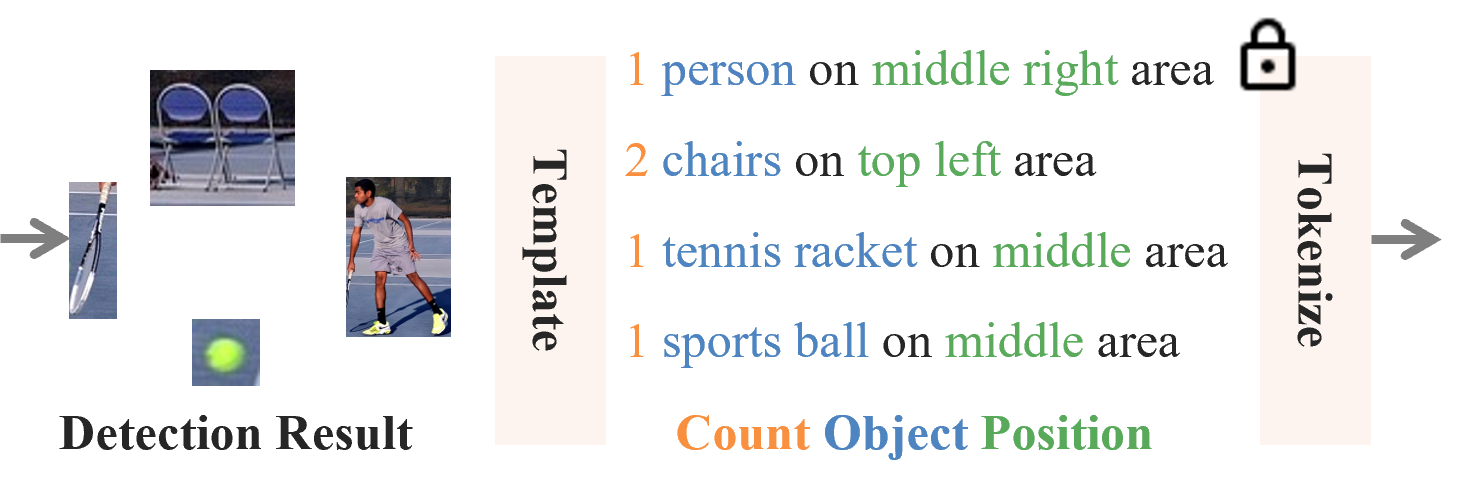}
    \begin{minipage}{0.9\linewidth}
      \caption{\textbf{Perception Forward Block} }
      \label{fig:imageb}
    \end{minipage}
  \end{subfigure}
  \begin{subfigure}[b]{0.5\textwidth}
    \centering
    \includegraphics[width=\linewidth]{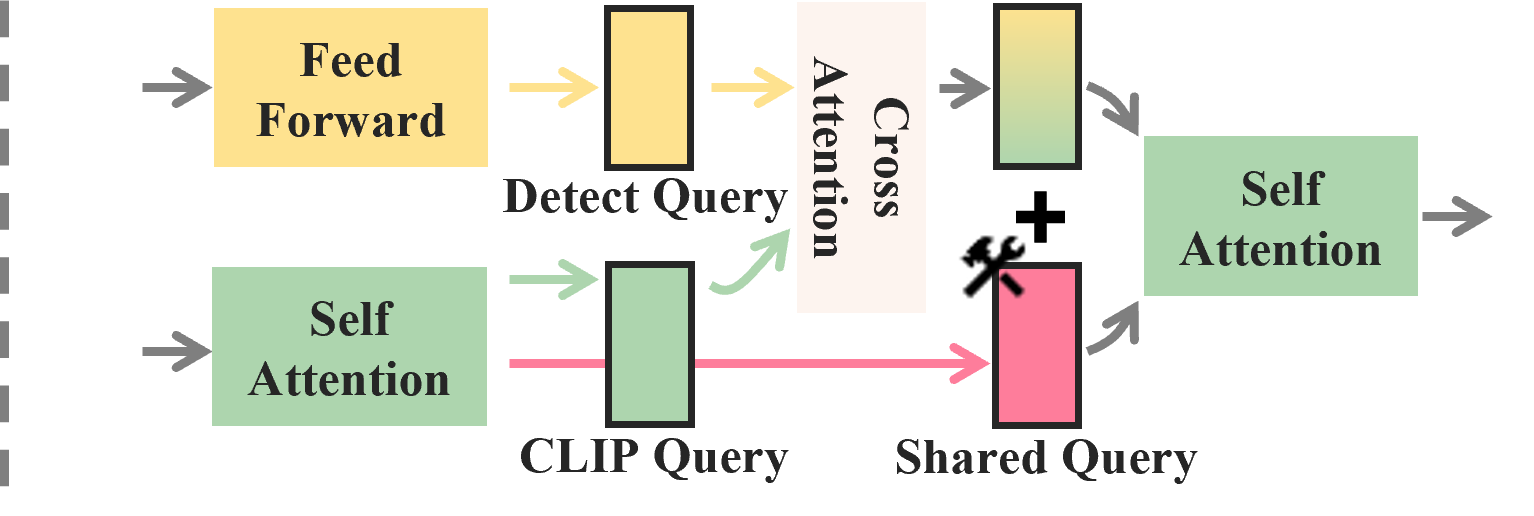}
    \begin{minipage}{0.9\linewidth}
      \caption{\textbf{Visual Forward Block} }
      \label{fig:imagec}
    \end{minipage}
  \end{subfigure}
  \caption{\textbf{Pipeline of MR-MLLM.} \textbf{(a) General Pipeline.} We employ a detection head encoder and a pre-trained CLIP encoder to extract object descriptors and scene descriptors from images, respectively. Queries containing semantic information at different scales from these encoders are aligned with adapter queries via an MLP layer and then added to them, infusing visual modality information into LLaMA. During the training process, the original parameters of LLaMA are frozen. \textbf{(b) Perception Forward Block.} The fine-grained object information output by the transformer-based object detection head is converted into a textual template, which is then tokenized into queries by the tokenizer of LLaMA. \textbf{(c) Visual Forward Block.} We introduce learnable shared queries to bridge the gap between visual perception and multimodal comprehension. Shared queries are updated during training.}
  \label{fig_pipeline}
\end{figure*}

\section{Methodology}
\label{sec:method}
In this section, we propose MR-MLLM for mutual reinforcement of multimodal comprehension and vision perception.
In section \ref{sec:overview}, we introduce the overall pipeline of MR-MLLM and provide an overview of its working mechanism and key components. 
Subsequently, in Section \ref{sec:SQF}, we elaborate on the shared query mechanism, emphasizing its crucial role in facilitating mutual reinforcement between vision perception and multimodal comprehension. 
In Section \ref{sec:PECMI}, we delve into perception-enhanced cross-modal integration, a pivotal innovation that introduces a perception-based modality to augment the multimodal learning process.
Finally, in Section \ref{sec:enhanced_prompt}, the perception-embedded prompt generation method is designed to innovatively integrate perceptual information into the prompt generation.

\subsection{Overall pipeline of MR-MLLM}
\label{sec:overview}

The proposed Mutually Reinforced Multimodal Large Language Model (MR-MLLM) framework, shown in Figure \ref{fig_pipeline}, represents a groundbreaking approach to multimodal learning. It integrates the strengths of visual perception and language comprehension, pioneering a shared query mechanism that synergistically enhances both tasks of multimodal comprehension and vision perception. This framework leverages the rich perceptual information from vision perception models and the contextual insights of large language models, harmonizing them to elevate multimodal comprehension.

Central to MR-MLLM is the dual-stream architecture, composed of a visual perception stream and a linguistic processing stream. The visual stream, powered by pre-trained visual encoders, \(f_V\), transforms raw images \(I\) into visual feature spaces \(V_f\):
\begin{equation}
\label{eq:3_7}
    H^{(n+1)} = \mathcal{L}(H^{(n)}, V_{proj} + P_{proj})
\end{equation}

In parallel, the linguistic stream, realized through a Large Language Model, \(f_L\), processes textual input \(T\) into rich contextual embeddings \(L_e\), adept at capturing the nuanced interdependencies within the text:
\begin{equation}
\label{eq:3_2}
    L_e = f_L(T)
\end{equation}

We introduce a novel shared query structure, \(Q_{shared}\), which is formulated by integrating the outputs from two pre-trained visual encoders, which are perception forward and visual forward. This integration, achieved through a sophisticated fusion function \(\mathcal{F}\), leverages multi-knowledge interaction to enhance mutual understanding:
\begin{equation}
\label{eq:3_3}
    Q_{shared} = \mathcal{F}(V_{fm}), V_{fp})
\end{equation}

Furthermore, MR-MLLM innovatively incorporates perception outputs as a new modality, enriching multimodal interactions. This is realized by embedding these perception outputs, \(P_o\), into the existing multimodal framework \(\mathcal{G}\):
\begin{equation}
    C_{MLLM} = \mathcal{G}(Q_{shared}, P_o)
\end{equation}

Additionally, MR-MLLM enhances traditional prompt generation by integrating perception-enhanced prompts into the Large Language Model. This integration leads to a more informed and contextually aware language model, significantly enhancing its response generation capabilities. The enhanced prompt generation manner is represented as \(E_{prompt}\), which refines the model's ability to generate relevant and contextually rich responses. This is mathematically articulated as:
\begin{equation}
\label{eq:3_5}
    E_{prompt} = \mathcal{P}(P_{o}, T_{orig})
\end{equation}

In this equation, \(\mathcal{P}\) represents the function responsible for encoding the enhanced prompts, \(P_{o}\) denotes the perception outputs integrated into the prompt, and \(T_{orig}\) is the original textual prompt. This innovative approach to prompt generation in MR-MLLM allows for a more dynamic interaction with the input data, leveraging both the rich perceptual information and the original textual context to produce a more accurate and comprehensive understanding of the given scenario.

In summary, the overall pipeline of MR-MLLM can be formulated as below:
\begin{equation}
MR\text{-}MLLM(I, T) = LLM\left( E(P_o, T_{\text{orig}}), \mathcal{G}(Q_{\text{shared}}, P_o) \right)
\end{equation}

In the proposed equation, \( MR\text{-}MLLM(I, T) \) delineates the process of our model handling the inputs: image \( I \) and text \( T \). The encoder \( E \) integrates the perception output \( P_o \) with the original text \( T_{\text{orig}} \), creating an enhanced prompt that enriches the contextual understanding for the Large Language Model. This procedure, \( E(P_o, T_{\text{orig}}) \), is pivotal in embedding perceptual insights into the linguistic domain. Concurrently, the function \( \mathcal{G} \) merges the shared query \( Q_{\text{shared}} \) with the perception output \( P_o \), facilitating a sophisticated cross-modal integration. This integration is instrumental in harmonizing the insights from both visual and textual modalities, thereby equipping the LLM with a comprehensive and nuanced understanding of the multimodal inputs. 
Furthermore, through the implementation of MR-MLLM, we can transcend the traditional paradigms by unifying the output modalities of both vision perception tasks and multimodal comprehension tasks into a singular language modality. This innovative approach facilitates the infusion of the LLM's extensive generalization and emergent capabilities into the vision perception models.
Such a mechanism exemplifies the innovative approach of MR-MLLM in leveraging both perceptual and linguistic elements to enhance multimodal learning and comprehension.


\subsection{Shared Query Fusion in Multimodal Learning}
\label{sec:SQF}

We introduce an innovative shared query fusion mechanism, a cornerstone in bridging the gap between visual perception and multimodal comprehension. This fusion is pivotal for synthesizing a holistic understanding from multimodal inputs.

Central to this fusion process is the integration of two distinct query streams. The first stream (visual forward), shown in Figure \ref{fig:imagec}, is derived from the visual aspect of the MLLM, where image features, extracted and transformed by a pre-trained image encoder, are projected into a specific query space. The second stream emanates from a pre-trained vision perception model (perception forward), contributing an additional layer of visual insights.

These two query streams are then intricately fused using a series of transformer blocks designed for cross-modal interactions. This fusion generates a shared query, \(Q_{shared}\), which encapsulates the combined strengths of both visual perception and linguistic analysis. The shared query is formulated as follows:
\begin{equation}
\label{eq:3_6}
    Q_{shared} = \mathcal{F}(f_V(I), f_P(P))
\end{equation}

In this equation, \(f_V(I)\) signifies the visual query derived from the image encoder, while \(f_P(P)\) indicates the query from the vision perception model. The fusion function \(\mathcal{F}\) encapsulates the sophisticated integration process within the transformer architecture, creating a unified query that encapsulates the attributes of both modalities.

The shared query \(Q_{shared}\) serves as the linchpin in MR-MLLM, facilitating a synergistic interplay between the vision perception and multimodal comprehension tasks. It enables the model to not only extract richer information from MLLM but also to leverage perception information for enhanced multimodal comprehension. This process epitomizes the mutual reinforcement concept, where each modality's strengths are harnessed and amplified, leading to a more robust and nuanced understanding of multimodal comprehension and perception.

In essence, the shared query fusion mechanism in MR-MLLM sets a new paradigm in multimodal learning, showcasing how the convergence of advanced vision perception and multimodal comprehension informed insights can yield superior performance.

\subsection{Perception-Enhanced Cross-Modal Integration}
\label{sec:PECMI}

In MR-MLLM, we introduce a pioneering approach that incorporates perception-enhanced cross-modal integration to significantly advance multimodal learning. This method fundamentally enriches the interplay between visual and linguistic modalities by introducing a third, perception-based modality derived from vision perception models.

\begin{itemize}
    \item \textbf{Incorporation of Perception Modality}: As shown in Figure \ref{fig:imageb}, our model innovatively incorporates perception data, such as bounding box outputs from object detection models, as an independent modality. This integration allows for the capture of intricate visual details often overlooked in standard image-text pair encodings. The perception modality (\(P\)) is represented as:
        \begin{equation}
            P = f_{\text{perception}}(\text{Visual Data})
        \end{equation}
        where \(f_{\text{perception}}\) denotes the function applied by the vision perception model on the visual data.

    \item \textbf{Advanced Cross-Modal Fusion}: As shown in Figure \ref{fig:imagec}, MR-MLLM leverages a novel cross-modal fusion mechanism where the perception modality \(P\) is synergized with the image modality (\(I\)) prior to the cross-attention phase with the language modality (\(L\)). This fusion process is mathematically represented as:
        \begin{equation}
            I_P = \text{fusion}(I, P)
        \end{equation}
        where \(I_P\) denotes the integrated image-perception representation.

    \item \textbf{Enhanced Multimodal Interaction}: The integrated representation \(I_P\) is then utilized in the cross-attention mechanism with the linguistic stream, enriching the overall multimodal understanding. This step effectively consolidates detailed visual cues with contextual linguistic information, leading to a deeper comprehension of multimodal content.
        \begin{equation}
            M = \text{cross\_attention}(I_P, L)
        \end{equation}
        where \(M\) represents the final multimodal comprehension.
\end{itemize}

\subsection{Perception-Embedded Prompt Generation}
\label{sec:enhanced_prompt}

We innovatively propose the integration of perceptual information into the prompt generation. Our MR-MLLM exemplifies this by embedding perception outputs into the prompt formulation, thereby enriching the language model's contextual awareness and response generation capabilities.

The essence of this process lies in transmuting perception outputs, which we denote as \( P_o \), into an enriched prompt space. This is achieved through a specialized encoder, represented by the function \( E \). The perception outputs, alongside the original prompt \( T \), undergo a transformation to yield an enhanced prompt \( E_{prompt} \), which is then fed into the language model. This process can be mathematically described as:

\begin{equation}
    E_{prompt} = E(P_o, T)
\end{equation}

In MR-MLLM, \( P_o \) encapsulates high-level perceptual information derived from the vision model, providing a rich contextual backdrop for the language model. By integrating \( P_o \) with the original textual prompt \( T \), the model achieves a more nuanced understanding of the multimodal input, thus generating responses that are not only contextually relevant but also perceptually informed.


\begin{table}[b]
\centering
\caption{
\textbf{Performance comparison for general VQA.} We establish a new state-of-the-art performance on the VSR dataset, which primarily evaluates spatial reasoning capabilities.}
\label{tab:vqa}
\resizebox{0.46\textwidth}{!}{%
\begin{tabular}{l|llll}
\toprule
Method & OKVQA & VQAV2 & VSR \\
\hline
BLIP2~\citep{li2023blip}                   & 45.9   & -     & 50.9 \\
Instruct BLIP~\citep{Dai2023InstructBLIPTG}           & -      & -     & 52.1 \\
LLaMA-AdapterV2~\citep{gao2023llamaadapter}         & 49.6   & 70.7  & -   \\
Shikra~\citep{chen2023shikra}                  & 47.2   & 77.4  & -   \\
Fuyu-8B~\citep{fuyu-8b}                 & 60.6   & 74.2  & -   \\
MiniGPT-v2~\citep{chen2023minigpt}              & 57.8   & -     & 62.9 \\
Qwen-VL-7B~\citep{Bai2023QwenVLAF}              & 58.6   & 79.5  & 63.8 \\
Qwen-VL-7B-Chat~\citep{Bai2023QwenVLAF}         & 56.6   & 78.2  & 61.5 \\
LLaVA1.5~\citep{Liu2023ImprovedBW}                   & -      & 78.5  & -   \\
\hline
\rowcolor{lightgreen}
OURS                    & 57.3   & 74.9  & \textbf{71.5} \\ 
\bottomrule                 
\end{tabular}%
}
\end{table}

\section{Experiments}
In this section, we conduct extensive experiments to validate the performance of the proposed MR-MLLM. In Section \ref{4_1}, we describe the experimental setups, which include the datasets employed, baseline models, evaluation metrics, and the implementation details of our approach. Subsequently, in Section \ref{4_2} and \ref{4_3}, we evaluate and analyze the performance of MR-MLLM on multimodal comprehension tasks and vision perception tasks with state-of-the-art MLLMs, respectively. Finally, in Section \ref{4_4}, we conduct ablation studies on MR-MLLM to verify the effectiveness of our proposed key perception interaction modules.

\label{sec:exp}
\newcommand{\yichi}[1]{\textcolor{magenta}{[yichi: #1]}}
\begin{table*}[htbp]
\centering
\caption{\textbf{Object hallucination benchmark using POPE evaluation pipeline}\cite{li2023evaluating}. Our model demonstrates a remarkable performance in discerning hallucinations in images with the smallest parameter scale listed in the table (7B). }
\label{tab:pope}
\begin{tabular}{c|c|ccccccc}
\bottomrule
Datasets                     & Metrics                        & \cellcolor{lightgreen} OURS            & Shikra & InstructBLIP & MiniGPT-4 & LLaVA & MM-GPT & mPLUG-Owl \\ \hline
\multirow{5}{*}{Popular}     & \multicolumn{1}{l|}{Accuracy}  & \cellcolor{lightgreen}\textbf{84.33} & 83.97  & 82.77        & 79.67     & 50.37 & 50.00  & 50.90     \\
                             & \multicolumn{1}{l|}{Precision} & \cellcolor{lightgreen}85.66          & 87.55  & 76.27        & 78.24     & 50.19 & 50.00  & 50.46     \\
                             & \multicolumn{1}{l|}{Recall}    & \cellcolor{lightgreen}82.47          & 79.20  & 95.13        & 82.20     & 99.13 & 100.00 & 99.40     \\
                             & \multicolumn{1}{l|}{F1-Score}  & \cellcolor{lightgreen}84.04          & 83.16  & 84.66        & 80.17     & 66.64 & 66.67  & 66.94     \\
                             & \multicolumn{1}{l|}{Yes Ratio} & \cellcolor{lightgreen}48.13          & 45.23  & 62.37        & 52.53     & 98.77 & 100.00 & 98.57     \\ \hline
\multirow{5}{*}{Random}      & \multicolumn{1}{l|}{Accuracy}  & \cellcolor{lightgreen}\textbf{89.00} & 86.90  & 88.57        & 79.67     & 50.37 & 50.10  & 53.97     \\
                             & \multicolumn{1}{l|}{Precision} & \cellcolor{lightgreen}94.59          & 94.40  & 84.09        & 78.24     & 50.19 & 50.05  & 52.07     \\
                             & \multicolumn{1}{l|}{Recall}    & \cellcolor{lightgreen}82.73          & 79.27  & 95.13        & 82.20     & 99.13 & 100.00 & 99.60     \\
                             & \multicolumn{1}{l|}{F1-Score}  & \cellcolor{lightgreen}88.26          & 86.19  & 89.27        & 80.17     & 66.64 & 66.71  & 68.39     \\
                             & \multicolumn{1}{l|}{Yes Ratio} & \cellcolor{lightgreen}43.73          & 43.26  & 56.57        & 52.53     & 98.77 & 99.90  & 95.63     \\ \hline
\multirow{5}{*}{Adversarial} & \multicolumn{1}{l|}{Accuracy}  & \cellcolor{lightgreen}80.87          & \textbf{83.10}  & 72.10        & 65.17     & 49.70 & 50.00  & 50.67     \\
                             & \multicolumn{1}{l|}{Precision} & \cellcolor{lightgreen}79.91          & 85.60  & 65.13        & 61.19     & 49.85 & 50.00  & 50.34     \\
                             & \multicolumn{1}{l|}{Recall}    & \cellcolor{lightgreen}82.47          & 79.60  & 95.13        & 82.93     & 99.07 & 100.00 & 99.33     \\
                             & \multicolumn{1}{l|}{F1-Score}  & \cellcolor{lightgreen}81.17          & 82.49  & 77.32        & 70.42     & 66.32 & 66.67  & 66.82     \\
                             & \multicolumn{1}{l|}{Yes Ratio} & \cellcolor{lightgreen}51.60          & 46.50  & 73.03        & 67.77     & 99.37 & 100.00 & 98.67     \\ \bottomrule
\end{tabular}

\end{table*}
\begin{table*}[htbp]
\centering
\caption{\textbf{Comparisons on NoCaps\cite{Agrawal_2019}.}  While most MLLMs\cite{li2023blip,Li2022BLIPBL, Dai2023InstructBLIPTG, chen2023shikra} tested on NoCaps require pre-training on other extensive datasets, MR-MLLM only fine-tune on COCO Caption~\citep{cocodataset}.}
\label{tab:caption}
\begin{tabular}{l|lllllll}
\hline
\multirow{2}{*}{Model} & \multicolumn{2}{l}{In-Domain} & \multicolumn{2}{l}{Near-Domain} & \multicolumn{2}{l}{Out-Domain} & Overall \\
                       & BLEU@4         & CIDEr        & BLEU@4          & CIDEr         & BLEU@4         & CIDEr         & CIDEr   \\ \hline
LLaMA-Adapter V2~\citep{gao2023llamaadapter}        & 40.9           & 74.6         & 40.4            & 77.6          & 31.1           & 77.8          & 74.7    \\ \hline
\rowcolor{lightgreen} OURS & \textbf{42.9} & \textbf{79.0} & \textbf{41.8} & \textbf{78.4} & \textbf{32.2} & 76.3 & \textbf{79.2}  \\ \bottomrule 
\end{tabular}
\end{table*}

\begin{table}[b]
\centering
\caption{\textbf{Detection results on the publicly available CODA dataset \cite{li2022coda}.} As shown in the table, all the experimental methods used SODA10M~\citep{han2021soda10m} as training data, and the performance was evaluated on the CODA base dataset. Following CODA, the experiments use mAR, $AR^{10}$ as the evaluation metric. \textbf{/} indicates that closed-set detection models are unable to detect novel objects.}
\label{tab:detection_tab}
\resizebox{0.48\textwidth}{!}{
\begin{tabular}{lcccccc}
\toprule
CODA \cite{li2022coda}    & \multicolumn{2}{c}{Corner} & \multicolumn{2}{c}{Common} & \multicolumn{2}{c}{Novel} \\
\midrule
Method                    & mAR & $AR^{10}$ & mAR & $AR^{10}$ & mAR & $AR^{10}$ \\
\midrule
RetinaNet~\citep{lin2018focal}        & \textbf{11.9} & 5.4 & 28.7 & 23.9 & / & / \\
Faster R-CNN~\citep{fastrcnn}         & 6.8           & 4.9 & 23.9 & 23.1 & / & / \\
Cascade R-CNN~\citep{DBLP:journals/corr/abs-1712-00726}        & 8.3 & 5.5 & 27.2 & 25.3 & / & / \\
ORE~\citep{9578456}                   & 8.3 & 5.6 & 18.5 & 18.1 & 3.4 & 2.9 \\
\rowcolor{lightgreen}
DETR~\citep{DETR}            & 5.5 & 3.5 & 35.8 & 31.6 & / & / \\
\rowcolor{lightgreen}
MR-DETR (OURS)                & 7.6 & \textbf{6.7} & \textbf{37.1} & \textbf{32.5} & 2.5 & 2.3 \\
\bottomrule
\end{tabular}
}
\end{table}
\begin{figure*}[t] 
\centering 
\includegraphics[width=\textwidth]{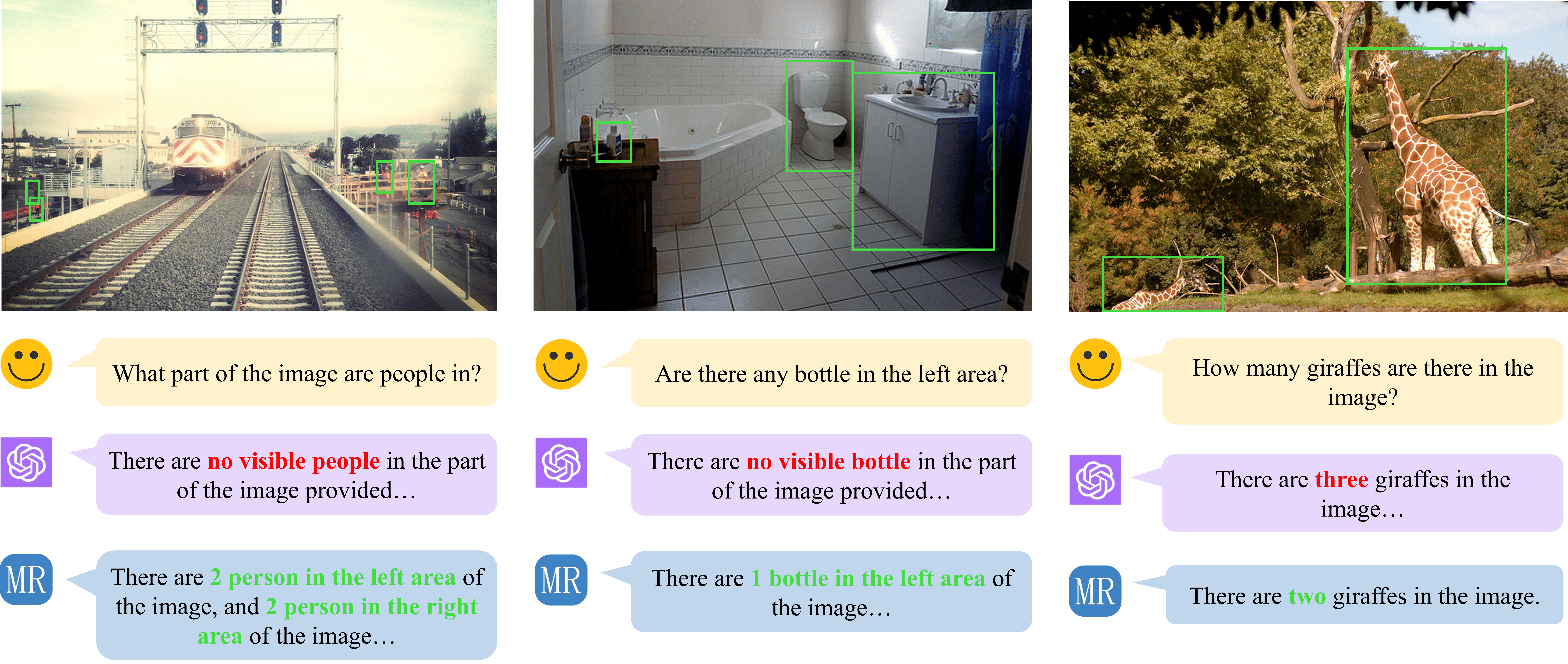} 
\caption{\textbf{MR-MLLM vs GPT-4V~\citep{gpt4llm}.} Due to the instance-level object descriptor provided by object detention head, MR-MLLM performs better than the mighty GPT-4V in some context involving spatial reasoning and fine-grained object perception.  } 
\label{vsgpt4} 
\end{figure*}
\subsection{Experimental Setups}
\label{4_1}
\textbf{Datasets.} We train MR-MLLM on 150K single-turn instruction data from LLaVA~\citep{llava}, and 860K object detection data in an instruction tuning format for the model's capability of refining coordinates. We also fine-tune  MR-MLLM on datasets for specific downstream tasks like VSR~\citep{Liu2022VisualSR}, VQA 2.0~\citep{Agrawal2015VQAVQ}, and COCO Caption~\citep{cocodataset}, etc.
\\
\textbf{Baseline methods.} 
In multimodal comprehension, MR-MLLM was tested against models like BLIP2~\citep{li2023blip}, Instruct BLIP~\citep{Dai2023InstructBLIPTG},LLaMA-Adapter V2~\citep{gao2023llamaadapter} and others, across VQA and captioning benchmarks. For vision perception, we evaluated it against models such as RetinaNet~\citep{lin2018focal} and DETR~\citep{DETR} using the CODA dataset.
\\
\textbf{Evaluation metrics.} 
To evaluate the reinforcement of multimodal comprehension, we employ VQA with accuracy as the metric and caption generation, using BLEU-4 and CIDEr scores for assessment. Additionally, we leverage the POPE pipeline to evaluate the model's level of Object hallucination. For the reinforcement of visual perception, we select object detection as our assessment task, mAR and $AR^{10}$ across various domains as metrics.
\\
\textbf{Implementation Details.} We reorganize all training data into instruction tuning format, in an order of single-turn conversations with one question and one answer. We use pre-trained CLIP as the visual encoder to extract scene descriptors, and a transformer-based detection model to extract object descriptors. We adopt simple MLPs as the projection layer before aligning different types of query. The normalization layers, linear layer bias and scale of the adapter, the projection layer to align different types of query, and the shared query in the visual forward module are updated during training, while the original parameters in LLaMA are frozen. All training is done on 8 A100 GPUs within an acceptable time. 

\subsection{Evaluation for the reinforcement of multimodal comprehension}
\label{4_2}
We evaluate general VQA benchmarks, such as OKVQA~\citep{Marino2019OKVQAAV}, VQA2~\citep{balanced_vqa_v2}, and visual spatial reasoning (VSR)~\citep{Liu2022VisualSR}. MR-MLLM has achieved a new state-of-the-art performance on the VSR dataset, reaching 71.5, which surpasses the previous SOTA~\citep{tan2019lxmert} by 1.4 points. The results are presented in Table~\ref{tab:vqa}. We also conducted evaluations on the A-OKVQA (Multiple Choice) dataset, where MR-MLLM achieved an accuracy of 66.8\%, significantly surpassing the 60.9\% achieved by the similarly adapter-based LLaMA-Adapter V2.

Furthermore, we conducted an evaluation of our method using the POPE~\citep{li2023evaluating} evaluation pipeline. POPE imposes stringent requirements on model's spatial reasoning capabilities and perception of fine-grained objects. The results are presented in Table~\ref{tab:pope}. Our model achieves performance comparable to  Shikra and surpass many  MLLMs with larger parameter size. Figure.~\ref{vsgpt4} presents examples of VQA that demonstrate the remarkable multimodal comprehension capabilities of MR-MLLM.

As for image caption ability, we choose NoCaps~\citep{Agrawal_2019} as the benchmark, which necessitates the model's learning of visual concepts. Most MLLMs require pre-training on extensive datasets for captioning tasks. For instance, BLIP~\citep{li2023blip} and BLIP2~\citep{li2023blip2} need pre-training on datasets such as COCO Caption~\citep{cocodataset}, Visual genome~\citep{krishna2017visual}, Conceptual Captions~\citep{changpinyo2021conceptual}, and LAION~\citep{laioncoco}, whose scale incurs significant training costs. In contrast, Clipcap~\citep{mokady2021clipcap}, LLaMA-Adapter V2~\citep{gao2023llamaadapter} and our model only requires fine-tuning on the COCO Caption dataset. MR-MLLM achieved an overall CIDEr score of 79.2, significantly outperforming ClipCap's 65.8 and LLaMA-Adapter V2's 74.7. Table~\ref{tab:caption} presents the performance across different domains in NoCaps captioning task.

\subsection{Evaluation for the reinforcement of vision perception}
\label{4_3}
To evaluate the enhancement of vision perception in our framework, we use DETR~\citep{DETR} as the perception model in MR-MLLM, hereafter referred to as MR-DETR. The experimental results, as depicted in Table~\ref{tab:detection_tab}, demonstrate the efficacy of our proposed MR-DETR method when trained on the SODA10M dataset and evaluated on the CODA dataset. Notably, MR-DETR outperforms the conventional DETR model and other baseline methods in various scenarios. In the context of the corner case, MR-DETR achieves a significant improvement in $AR^{10}$, registering a score of 6.7\% compared to DETR's 3.5\%, illustrating its robustness in challenging scenarios. Furthermore, in the common case, MR-DETR attains the highest mAR of 37.1\% and $AR^{10}$ of 32.5\%, surpassing the DETR's scores of 35.8\% and 31.6\%, respectively. These results underscore the superiority of MR-DETR in handling common scenarios with higher accuracy and efficiency. Although the performance in the novel case indicates room for further improvement, the overall results validate the effectiveness of MR-DETR in generalizing the Large Language Model's capabilities to enhance vision perception tasks, thereby advancing the state-of-the-art in object detection under diverse and challenging conditions.
\begin{figure}[t] 
\centering 
\includegraphics[width=0.45\textwidth]{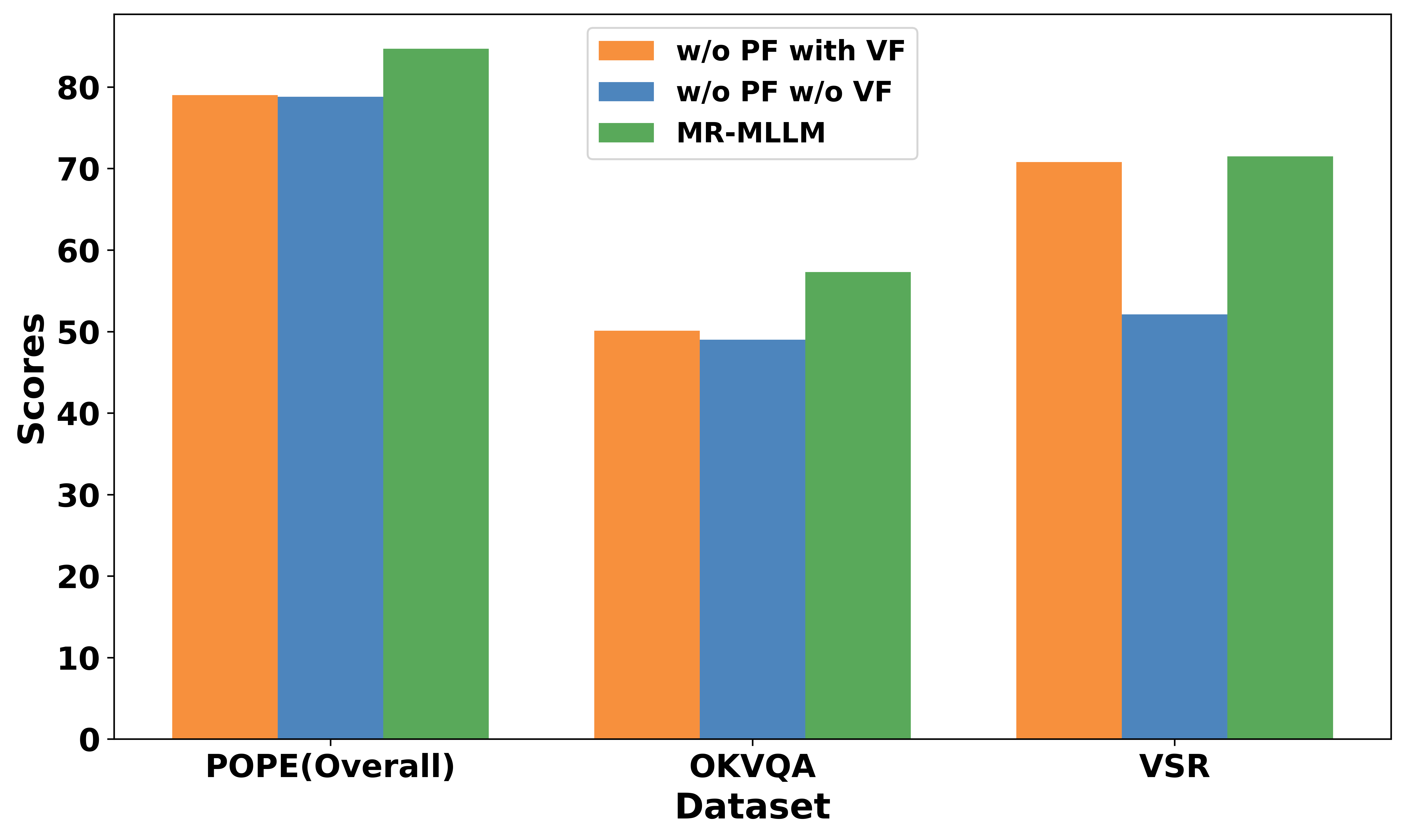} 
\caption{\textbf{Results from the ablation study.}The graph compares the baseline model (without modules) to variations where one or both modules (\textbf{PF} stands for Perception Forward Module while \textbf{VF} stands for Visual Forward Module) are incorporated, highlighting their contributions to the system's overall effectiveness.}
\label{Fig:ablation} 
\end{figure}
\subsection{Ablation Studies}
\label{4_4}


The Perception Forward module is designed to convert the detection results from detection head into sentence templates, which are then tokenized and injected directly into the LLM as textual information. Besides, The Visual Forward module significantly contributes to enhancing the model's attention to details derived from the detection head. Through the introduction of shared queries, this module empowers the model to intricately focus on and process the fine-grained intricacies presented by the detection head.

We conducted ablation studies to evaluate the effectiveness of these two modules.When both the Perception Forward module and Visual Forward module are removed, output by the detection head and queries from the CLIP Encoder are directly aligned with the adapter's queries through a projection layer and then added together. The discernible outcomes of these modifications and their impact on the system's performance are clearly depicted in Figure.~\ref{Fig:ablation}.

\section{Conclusion}
\label{sec:conclusion}

This study presents the Mutually Reinforced Multimodal Large Language Model (MR-MLLM), a framework that synergistically enhances vision perception and multimodal comprehension. MR-MLLM combines MLLMs' generalization with vision models' detailed perception, bridging multimodal comprehension and visual understanding gaps. Our extensive experiments on benchmarks for visual question answering, image captioning, and object detection showcase MR-MLLM's exceptional performance, particularly in fine-grained multimodal comprehension tasks and corner case vision perception. These results position MR-MLLM as a significant advancement in multimodal learning, offering new prospects for multimodal AI systems.

{
    \small
    \bibliographystyle{unsrtnat}
    \bibliography{main}
}

\clearpage
\setcounter{page}{1}
\maketitlesupplementary
\appendix

\section{Overview}
To enhance the comprehensiveness of our experiments, we have further elaborated on the MR-MLLM's performance in multimodal comprehension and vision perception in the supplementary material. The additional content is presented from the following perspectives.

\begin{itemize}
    \item Supplementary Experimental Analysis
    
    \noindent\hspace*{0.2em}- Multimodal comprehension with Grounding DINO

    \noindent\hspace*{0.2em}- Vision perception with Grounding DINO
    
    \item Additional Visualization Results
    

    \noindent\hspace*{0.2em}- Visualization results of image captioning
    
    \noindent\hspace*{0.2em}- Visualization results of visual understanding
    
    \item Expanded Related Work
    
    \noindent\hspace*{0.2em}- Expanded related work in Object Detection

    \noindent\hspace*{0.2em}- Expanded related work in Position Representation
    
    \item \href{https://youtu.be/X1-I-A2F-Ig}{Demo Video}

\end{itemize}

\section{Supplementary Experimental Analysis}
To validate the reinforcement of visual perception capabilities in other vision perception models provided by our proposed MR-MLLM, as well as the enhancement of multimodal comprehension abilities in MLLMs when integrating additional vision perception heads, we utilized Grounding DINO (GR-DINO) as the foundational perception model. This model was integrated into the MR-MLLM framework, leading to the development of a new MR-MLLM structure. We conducted experimental validations on several VQA datasets and object detection datasets to confirm these enhancements.
\subsection{Multimodal comprehension with GR-DINO}
To further explore the extensibility of our proposed MR-MLLM, we utilized Grounding DINO as the base vision perception model, combined with LLAMA, to construct the MR-MLLM. Subsequently, to verify the enhancement in multimodal comprehension capabilities of MLLM by the MR-MLLM structure, we conducted experiments on a series of visual question answering benchmarks compared with the base MLLM (LLaMA-adapter v2), with results presented in Table \ref{supply_vqa_tab}. It is evident that the new MLLM structure developed using MR-MLLM achieves higher VQA accuracy with equivalent data fine-tuning compared to the base MLLM on most datasets. 
Additionally, it was observed that after substituting DETR with Grounding DINO, performance on most benchmarks still surpassed the base MLLM, albeit with a slight decline compared to incorporating DETR as the base vision perception model. This can be attributed to the fact that Grounding DINO's queries not only encompass vision information but also text information, constituting a type of multimodal query. Hence, within our structure emphasizing fine-grained visual perception queries, the performance is not as optimal as MR-MLLM with DETR, yet it is better than the base MLLM.
These experimental outcomes prove that our proposed architecture, which integrates the vision perception model with MLLM, improves the multimodal comprehension capabilities of MLLM. This underscores the robust extensibility of our MR-MLLM and demonstrates its potential for plug-and-play applications.

Furthermore, to validate the fine-grained multimodal comprehension capabilities of our MR-MLLM in more challenging scenarios, we conducted a series of experiments on the POPE dataset using different detection heads as the base vision perception models. The experimental results are presented in Table \ref{supply_pope}. Compared to the conventional MLLM structure, our proposed MR-MLLM, which integrates vision perception with multimodal large language models, achieves a fine-grained comprehension of multimodal information.
\begin{table*}[htbp]
\centering
\renewcommand{\arraystretch}{1.5} 
\caption{\textbf{Results of MR-MLLM with different Detection Heads on various VQA Datasets.}}
\label{supply_vqa_tab}
\begin{tabular}{c|clllccc}
\hline
\multirow{3}{*}{Detection Head}     & \multicolumn{4}{c}{A-OKVQA}                                                                           & \multirow{3}{*}{OKVQA}   & \multirow{3}{*}{VQAV2}   & \multirow{3}{*}{VSR}     \\
                                    & \multicolumn{2}{c}{Multiple Choice}              & \multicolumn{2}{c}{Direct Answer}                  &                          &                          &                          \\
                                    &  \multicolumn{1}{c}{Val}                   & \multicolumn{1}{c}{Test} & \multicolumn{1}{c}{Val} & \multicolumn{1}{c}{Test} &                          &                          &                          \\ \hline

\multicolumn{1}{c|}{LLaMA Adapter V2}           & \multicolumn{1}{c}{-} & \multicolumn{1}{c}{60.90}                    & -                       & 48.85                    & \multicolumn{1}{c}{49.6} & \multicolumn{1}{c}{70.7} & \multicolumn{1}{c}{-} \\ \hline
\rowcolor{green!5}
\multicolumn{1}{c|}{OURS with DETR}           & \multicolumn{1}{c}{68.24} & \multicolumn{1}{c}{66.76}                    & 54.68                       & 53.63                    & \multicolumn{1}{c}{57.3} & \multicolumn{1}{c}{74.9} & \multicolumn{1}{c}{71.5} \\ \hline
\rowcolor{green!5}
\multicolumn{1}{l|}{OURS with GR-DINO} & \multicolumn{1}{c}{63.06} & \multicolumn{1}{c}{59.18}                        &   49.29                     & 49.55                        & \multicolumn{1}{c}{52.8}    & \multicolumn{1}{c}{70.9}    & \multicolumn{1}{l}{64.4}    \\ \hline
\end{tabular}
\end{table*}
\begin{table}[htbp]
\centering
\caption{\textbf{Object hallucination benchmark using POPE evaluation pipeline on MR-MLLM with different DEtection Heads.} GR-DINO stands for Grounding DINO.}
\label{supply_pope}
\begin{tabular}{c|lcc}
\hline
\multicolumn{1}{l|}{\multirow{2}{*}{Datasets}} & \multirow{2}{*}{Metrics} & \multirow{2}{*}{\begin{tabular}[c]{@{}c@{}}MR-MLLM\\ DETR\end{tabular}} & \multicolumn{1}{c}{\multirow{2}{*}{\begin{tabular}[c]{@{}c@{}}MR-MLLM\\ GR-DINO\end{tabular}}} \\
\multicolumn{1}{l|}{}                          &                          &                                                                         &                                                                                                   \\ \hline
\multirow{5}{*}{Popular}                       & Accuracy                 & 84.33                                                                   & \multicolumn{1}{c}{77.97}                                                                         \\
                                               & Precision                & 85.66                                                                   & \multicolumn{1}{c}{74.46}                                                                         \\
                                               & Recall                   & 82.47                                                                   & \multicolumn{1}{c}{85.13}                                                                         \\
                                               & F1-Score                 & 84.04                                                                   & \multicolumn{1}{c}{79.44}                                                                         \\
                                               & Yes Ratio                & 48.13                                                                   & \multicolumn{1}{c}{57.16}                                                                         \\ \hline
\multirow{5}{*}{Random}                        & Accuracy                 & 89.00                                                                   & \multicolumn{1}{c}{86.10}                                                                         \\
                                               & Precision                & 94.59                                                                   & \multicolumn{1}{c}{86.76}                                                                         \\
                                               & Recall                   & 82.73                                                                   & \multicolumn{1}{c}{85.20}                                                                         \\
                                               & F1-Score                 & 88.26                                                                   & \multicolumn{1}{c}{85.97}                                                                         \\
                                               & Yes Ratio                & 43.73                                                                   & \multicolumn{1}{c}{49.10}                                                                         \\ \hline
\multirow{5}{*}{Adversial}                     & Accuracy                 & 80.87                                                                   & \multicolumn{1}{c}{73.57}                                                                         \\
                                               & Precision                & 79.91                                                                   & \multicolumn{1}{c}{69.24}                                                                         \\
                                               & Recall                   & 82.47                                                                   & \multicolumn{1}{c}{84.80}                                                                         \\
                                               & F1-Score                 & 81.17                                                                   & \multicolumn{1}{c}{76.24}                                                                         \\
                                               & Yes Ratio                & 51.60                                                                   & \multicolumn{1}{c}{61.23}                                                                         \\ \hline
\end{tabular}

\end{table}

\subsection{Vision perception with GR-DINO}
To confirm the versatility of our MR-MLLM, we substituted the base vision model within the MR-MLLM structure from DETR (ResNet-50 backbone) to the open-vocabulary detector Grounding DINO (swin-T backbone)~\citep{liu2023grounding}. A series of experiments were conducted on the CODA dataset to evaluate the improved vision perception capability of MR-MLLM in corner cases. Experimental results are shown in Table \ref{tab:supply_detection_tab}.

Grounding DINO (GR-DINO), pre-trained on a vast corpus of image-text data, inherently possesses the capability to detect novel classes as an open-vocabulary detector. Its intrinsic performance on the CODA dataset significantly surpasses that of conventional vision detectors trained in a close-set manner. Following optimization with the MR-MLLM framework, MR-Grounding benefits from the compressed world knowledge of large MLLMs, exhibiting further enhancements in performance. This improvement is particularly notable in the detection capabilities for novel classes.

The aforementioned experiments further validate the extensibility of our MR-MLLM framework and its enhancement to the perception capabilities of vision perception models. In our subsequent research, we will continue to focus on the empowerment of domain-specific models through MLLMs.

\begin{table}[h]
\centering
\caption{\textbf{Detection results on the publicly available CODA dataset \cite{li2022coda}.} As shown in the table, all the experimental methods used SODA10M~\citep{han2021soda10m} as training data, and the performance was evaluated on the CODA base dataset. Following CODA, the experiments use mAR, $AR^{10}$ as the evaluation metric. \textbf{/} indicates that closed-set detection models are unable to detect novel objects.}
\label{tab:supply_detection_tab}
\resizebox{0.48\textwidth}{!}{
\begin{tabular}{lcccccc}
\toprule
CODA \cite{li2022coda}    & \multicolumn{2}{c}{Corner} & \multicolumn{2}{c}{Common} & \multicolumn{2}{c}{Novel} \\
\midrule
Method                                & mAR           & $AR^{10}$ & mAR  & $AR^{10}$ & mAR & $AR^{10}$ \\
\midrule
RetinaNet~\citep{lin2018focal}        & 11.9          & 5.4       & 28.7 & 23.9      & /   & / \\
Faster R-CNN~\citep{fastrcnn}         & 6.8           & 4.9       & 23.9 & 23.1      & /   & / \\
Cascade R-CNN~\citep{DBLP:journals/corr/abs-1712-00726}        
                                      & 8.3           & 5.5       & 27.2 & 25.3      & /   & / \\
ORE~\citep{9578456}                   & 8.3           & 5.6       & 18.5 & 18.1      & 3.4 & 2.9 \\
\rowcolor{lightgreen}
DETR~\citep{DETR}             & 5.5  & 3.5          & 35.8         & 31.6          & /         & /    \\
\rowcolor{lightgreen}
MR-DETR (OURS)                & 7.6  & 6.7          & 37.1         & 32.5          & 2.5       & 2.3  \\
\rowcolor{lightgreen}
Grounding DINO \cite{liu2023grounding}              
                              & 55.1 & \textbf{19.0}        & 58.4         & \textbf{51.6}          & 54.5      & 52.8  \\
\rowcolor{lightgreen}
MR-Grounding DINO (OURS)      & \textbf{56.1} & 18.7        & \textbf{58.5}         & 49.6          & \textbf{56.9}      & \textbf{53.1}  \\
\bottomrule
\end{tabular}
}
\end{table}
\section{Additional Visualization Results}
To visually demonstrate the mutual reinforcement of multimodal comprehension and vision perception by our proposed MR-MLLM, this section will present intuitive heatmaps and visualization results of MR-MLLM across various tasks.
\subsection{Visualization results of image captioning}
\begin{figure*}[t] 
\centering 
\includegraphics[width=\textwidth]{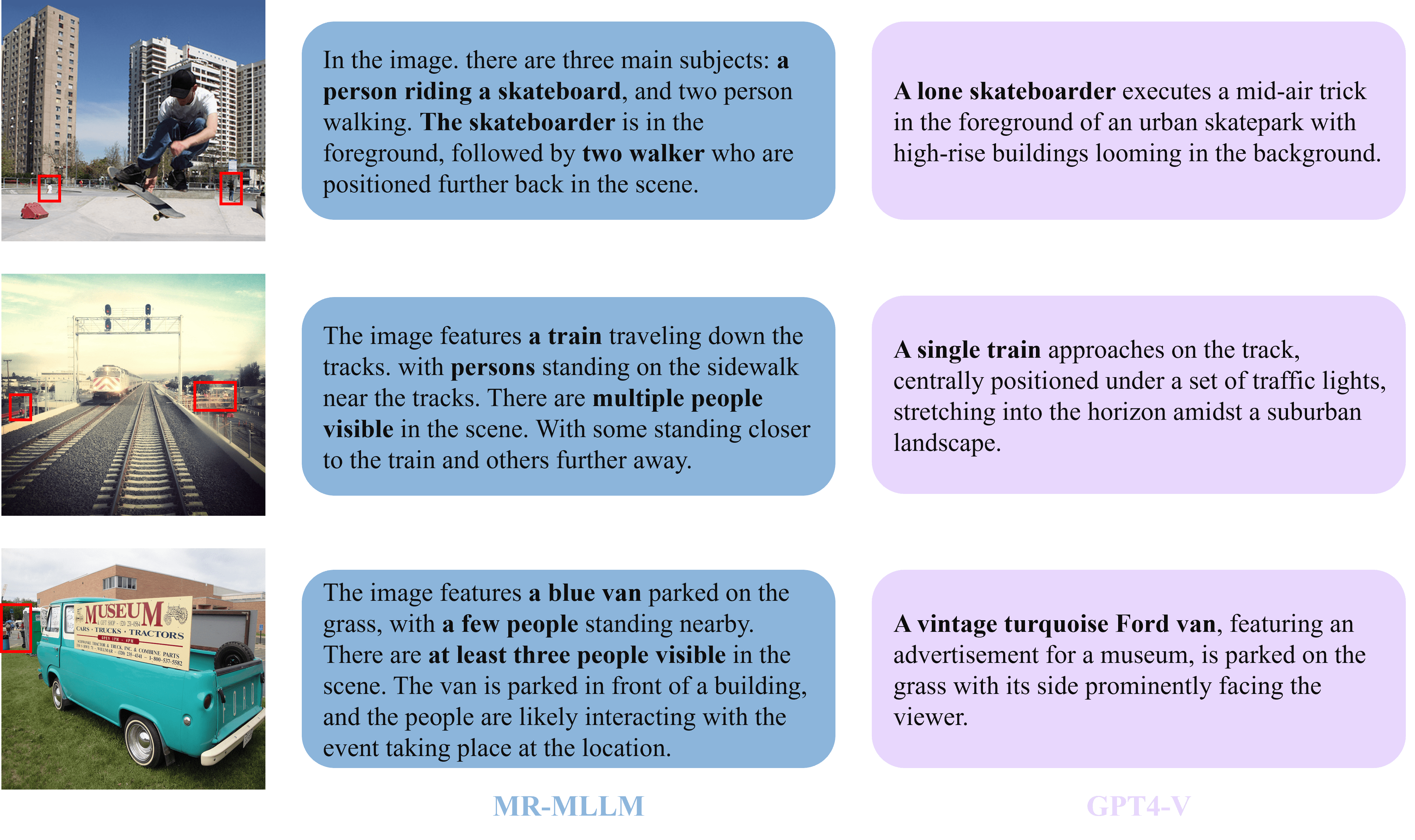} 
\caption{\textbf{Examples on Image Caption from MR-MLLM and GPT-4V~\citep{gpt4llm}.} Note that the red boxes on the image are not the output of detection heads; they are manually added to highlight the detailed objects to the reader.  } 
\label{sup_vsgpt4} 
\end{figure*}
Initially, we visually demonstrate the performance of MR-MLLM compared to GPT4-V in fine-grained image captioning tasks, as shown in Figure \ref{sup_vsgpt4}. Our method is capable of capturing intricate details of objects in images, providing captions with greater granularity. In contrast, general MLLMs tend to focus on macro-level captioning and often overlook instance-level informational nuances.

\subsection{Visualization results of high-level visual understanding}
\begin{figure*}[t] 
\centering 
\includegraphics[width=\textwidth]{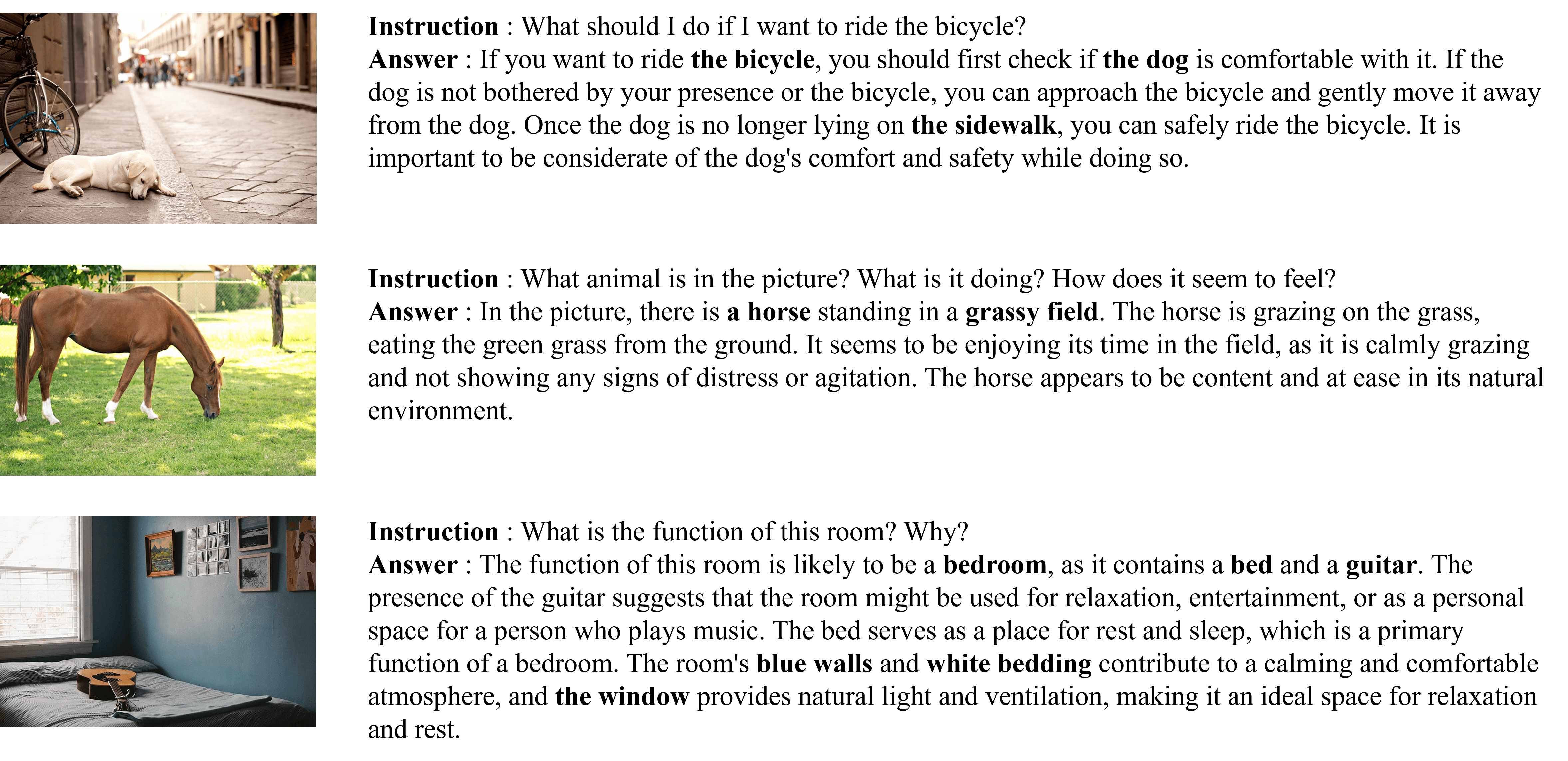} 
\caption{\textbf{More Visual Understanding Examples of MR-MLLM.}  After incorporating the ability to perceive fine-grained objects, MR-MLLM still demonstrates robust high-level visual understanding capabilities.} 
\label{sup_understanding} 
\end{figure*}
After incorporating the ability to perceive fine-grained objects, our method continues to demonstrate robust high-level visual understanding capabilities. Figure \ref{sup_understanding} illustrates some examples when our approach faces questions requiring multi-step generalized reasoning.

From the visualization examples in visual understanding tasks, it is apparent that our MR-MLLM pays more attention to fine-grained instances and their relative spatial relationships. It is able to extract deeper layers of perception information, thereby achieving a more detailed and comprehensive understanding of visual scenes at a finer granularity.

\end{document}